\documentclass[runningheads]{llncs}
\usepackage[T1]{fontenc}
\usepackage{graphicx}
\usepackage{numberedblock}
\usepackage{qtree}
\usepackage{tikz}
\usetikzlibrary{calc, trees}
\usepackage{caption}
\usepackage{enumitem}
\makeatletter
\newcommand\footnoteref[1]{\protected@xdef\@thefnmark{\ref{#1}}\@footnotemark}
\makeatother

\begin{document}
%
%
\title{\systemname{} : A Prolog Synergized Language Model for explainable Domain Specific Knowledge Based Question Answering}
%
%
\titlerunning{ProSLM}
%
%
\author{Priyesh Vakharia\inst{1} \and
Abigail Kufeldt\inst{1} \and
Max Meyers\inst{1} \and
Ian Lane\inst{1} \and
Leilani H. Gilpin\inst{1}}
%
\authorrunning{P. Vakharia et al.}
%
\institute{Department of Computer Science, School of Engineering,
University of California Santa Cruz, Santa Cruz, CA, USA, \email{\{pvakhari, akufeldt, mmeyers1, ialane, lgilpin\}@ucsc.edu}}
%
\index{Vakharia, Priyesh}
\index{Kufeldt, Abigail}
\index{Meyers, Max}
\index{Gilpin, Leilani H.}
%
\newcommand{\systemname}[1]{ProSLM}
\maketitle              
\begin{abstract}

Neurosymbolic approaches can add robustness to opaque neural systems by incorporating explainable symbolic representations. However, previous approaches have not used formal logic to contextualize queries to and validate outputs of large language models (LLMs). We propose \systemname{}, a novel neurosymbolic framework, to improve the robustness and reliability of LLMs in question-answering tasks. We provide \systemname{} with a domain-specific knowledge base, a logical reasoning system, and an integration to an existing LLM. This framework has two capabilities (1) context gathering: generating explainable and relevant context for a given query, and (2) validation: confirming and validating the factual accuracy of a statement in accordance with a knowledge base (KB). Our work opens a new area of neurosymbolic generative AI text validation and user personalization.

\end{abstract}
\section{Introduction}


Large language models (LLMs) are changing the way in which data is
used and information is disseminated \cite{genai_impacts}.  For
example, LLM-driven recommendation systems decide what content we view
\cite{rec_systs}.  Interactive question answering (QA) systems, e.g., OpenAI's ChatGPT, answers questions, or prompts,
in natural language text~\cite{chatgpt_overview}, leading to the
development of ``prompt engineering''\cite{white2023prompt} to guide the LLM on a
variety of tasks from questions in medical diagnosis~\cite{mesko2023prompt} to customer service~\cite{george2023review}.  Although LLMs are more than question-answering or
QA systems, they exhibit compelling emergent abilities: few-shot
prompting capabilities -- not observed in smaller models
\cite{emergent_abilities_llm}. Unsurprisingly, recent state of the art
LLMs like OpenAI's GPT or Meta's LLaMA models have exhibited
incredible performance and generalization across a multitude of
different tasks and domains, from telecommunications~\cite{wang2023network} to clean energy~\cite{rane2023contribution}
to healthcare~\cite{llm_survey_zhao}.

Despite their demonstrated power and social influence, these models are not trustworthy~\cite{wang2023decodingtrust}. LLMs, like many deep neural network, are considered ``black box'' systems; their large number of  parameters makes their reasoning uninterpretable to humans \cite{black_box}. Although LLMs can be prompted to self-explain \cite{predict_explain}, these explanations are unreliable~\cite{huang2023can}. Further, inaccurate model outputs, known as model hallucinations~\cite{hallucination-survey}, contribute to LLMs’ lack of reliability. We argue that LLMs need additional trustworthy explanations to empower users to both understand their line of reasoning and validate the accuracy and faithfulness of their outputs. Our goal is develop methods that use rules to detect imprecise textual statements with respect to a domain-specific knowledge base.

Symbolic AI, a subfield comprising of rule-based, deterministic techniques, can be used to generate trustworthy, human-interpretable explanations \cite{nesy_trends}. Symbolic approaches do not benefit from the robustness, creativity, and generalization ability exhibited by an LLM, but can be deployed in conjunction with a neural system to leverage the advantages of both techniques, also known as neurosymbolic AI. Symbolic components can introduce faithfulness and reliability into a neural system that cannot be achieved by an LLM alone \cite{nesy_survey}.

\begin{figure}[ht!]
\centering
\scalebox{0.85}{\includegraphics[width=\textwidth]{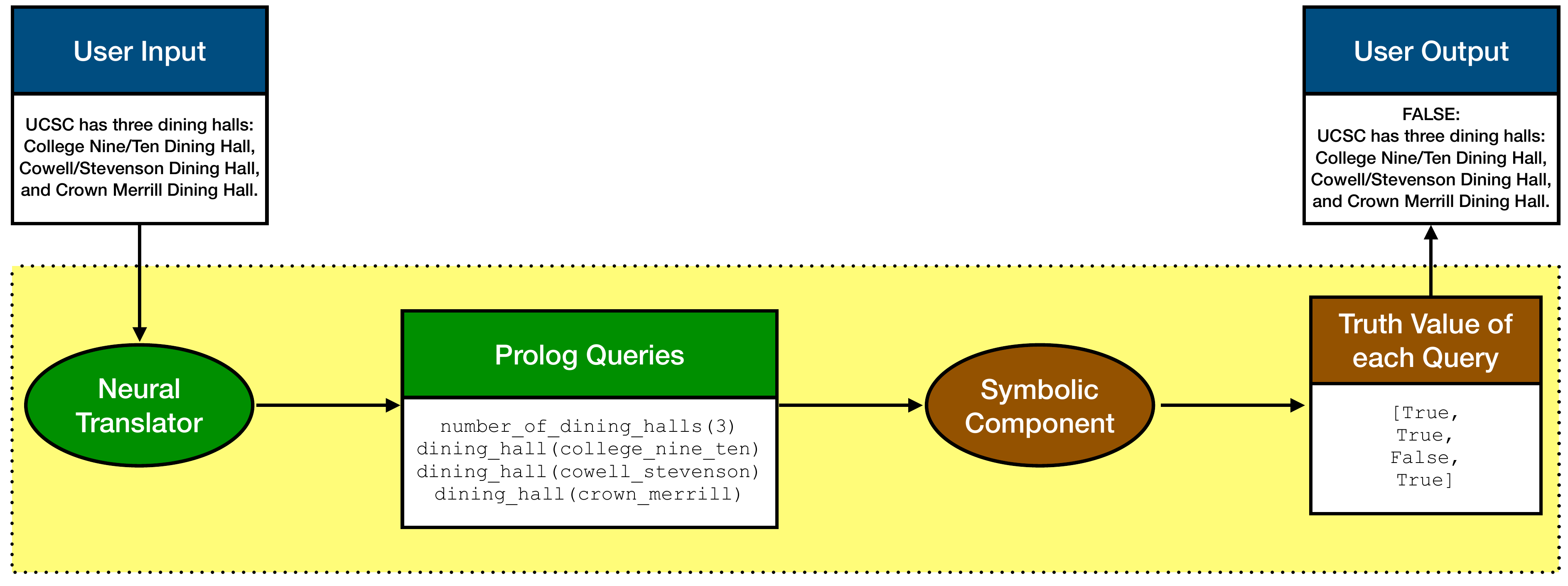}}
\caption{\small \systemname{} system workflow diagram}  \label{fig3}
\end{figure}

Toward this goal, we propose \systemname{}, as shown in Figure \ref{fig3}, a novel neurosymbolic framework designed to be a robust and trustworthy knowledge base question-answering (KBQA) system. \systemname{} introduces a symbolic component for \emph{explainable context gathering} prior to querying an LLM, using a knowledge base (KB) represented in Prolog. This symbolic component links a formal logic-based inference engine to a knowledge base of domain-specific information, and has two capabilities: (1) generating a retrievable and explainable chain of reasoning to contextualize the input query or (2) validating the accuracy of a given statement. With this, we leverage the creativity and generalizability of a pretrained LLM while improving the system’s robustness and explainability, empowering the user to better understand how the system came to a particular output, and in some cases identify model hallucinations or inaccuracies themselves. Further, this framework is compute-efficient, as the LLM does not require any additional training or finetuning for the domain of interest. 

\vspace{-2em}
\vspace{-2em}
The remainder of this paper is organized as follows. Section 2 overviews relevant prior work. Section 3 discusses the proposed framework. Experimental results are presented in Section 4, and analyzed in detail in Section 5. Lastly, Section 6 concludes the paper and discusses future directions for the continued development of this framework.

\section{Background}

Our work takes inspiration from previous neural and neurosymbolic methods aimed at improving model explainability.

\subsection{Hallucination}

Hallucinations in a language model refer to outputs that in some way diverge from factual truth \cite{hallucination_definition}. A recent example was when two attorneys used ChatGPT to draft a court filing and faced punishment after the model cited entirely fictitious court cases. This demonstrates that any trustworthy KBQA system needs validation and sanity checks. 

A new taxonomy of these errors proposed in \cite{hallucination_survey_huang} divides hallucinations into factuality and faithfulness hallucinations. The former encompasses both minor factual inconsistencies and complete fabrications. Faithfulness hallucinations refer to divergence from the prompt instruction, the context of the query, or logic itself -- prompt, context, and logic inconsistencies, respectively. 

As for their cause, hallucinations can originate in any part of the language modeling process, from data to the inference step \cite{hallucination_survey_huang}\cite{hallucination_origin}. Still, they can be detected and sometimes mitigated with the right strategies \cite{hallucination_mitigation}. When seeking to improve model explainability, then, it is prudent to strive to detect and mitigate hallucinations; in our case, \systemname{}'s symbolic component mitigates hallucinations by grounding the model with logical, relevant context, and has the potential to detect them by determining the truth value of a claim. This is discussed further in Section~\ref{fact-checking}. 

\subsection{Neural Approaches}

To alleviate the inherently uninterpretable nature of LLMs, many noteworthy neural approaches have been proposed. 

The PredictAndExplain method, introduced in \cite{predict_explain}, illustrates that models can be trained to generate explanations of their outputs without sacrificing performance, but this approach requires access to a reliable explanation-augmented dataset for training. 

Instead of training from scratch, chain of thought (CoT) prompting leverages large pre-trained models by instructing the model to describe the “chain of thought” it utilized to come to its answer during generation \cite{cot_prompting}. This method outperforms baselines in arithmetic, commonsense, and symbolic reasoning tasks, but requires sufficiently large LLMs in order to be successful. 

Retrieval augmented generation (RAG) architectures, first introduced in \cite{rag}, allows one to retrieve relevant documents from a database and input those texts as additional context to a language model. As opposed to CoT, this architecture directly feeds useful information to the QA model, thus grounding the response, reducing factuality hallucinations, and improving explainability of results. However, this approach becomes insufficient when given questions whose answers cannot be found in existing documents but rather must be deducted through logical inference and world knowledge. 


\subsection{Neurosymbolic Approaches}

Now we turn to neurosymbolic AI, a re-emergent field of techniques whose primary strengths include improved model transparency and explainability. 

One recent example of this in the visual question-answering (VQA) domain is the Neuro-Symbolic Concept Learner, a framework which jointly learns images, words, and semantic relationships from natural supervision using three modules -- a neural perception module, a semantic parser, and a symbolic program executor \cite{nscl}. As they demonstrate in their paper, the use of a DSL specifically designed for VQA to translate natural language queries into executable programs via formal semantics is powerful in its ability to generalize to new DSLs, queries, and images. In a similar fashion, \cite{logic_lm} proposed a framework to incorporate First-Order Logic with LLMs to solve logical reasoning tasks. 

Neither \cite{nscl} nor \cite{logic_lm} pose a viable approach in this current work, however, due to the complexity of questions expected for a KBQA system, and our subsequent interest in leveraging the generative power of an LLM to answer inputted queries. However, we can take inspiration from both approaches' use of an interpretable symbolic translator to generate useful context for the LLM to reason upon.


\section{Preliminaries}
\paragraph{First-Order Logic}
allows for a language representation using predicates, constants and functions. Predicate symbols represent relations on certain objects. Objects are represented by constant symbols. A \emph{term} refers to an object and can either use a function or not. Given that now we can represent objects and relations for those objects, we can now create sentences using logical connectives. We use atomic sentences, made of a predicate and objects, to state facts. An \emph{atom} is simply a fact. \texttt{friends(priyesh, abigail)} represents an atomic sentence that is interpreted as `priyesh and abigail are friends'. A \emph{literal} is an atom or its negation. A \emph{positive literal} is simply an atom. A definite clause is a clause with only one positive literal. For example, if \texttt{human(priyesh)}, \texttt{human(abigail)}, and \texttt{friends(priyesh, abigail)} are atoms, then:

\texttt{friends(priyesh, abigail) $\lor$ $\neg$ human(priyesh) $\lor$ $\neg$ human(abigail)} \hfill
is a definite clause and it is represented as 

\texttt{friends(priyesh, abigail) :- human(priyesh), human(abigail)}.

\paragraph{Backward Chaining}
is used by logical inference algorithms over definite clauses. These algorithms work backwards from the goal, chaining through rules to find known facts that support the proof \cite{AITextbook}. A goal is proved, and returns True, if the KB has a rule $lhs => goal$ where the lhs is made of conjuncts, with each conjunct an operation on two logical values, and if the lhs can be proved True. The algorithm then makes every conjunct the next goal, essentially running a recursive depth-first search for the first possible rule that returns True. 

\section{Method}

As previously discussed, LLMs lack trustworthy explainability~\cite{wang2023decodingtrust}. To alleviate this, we introduced \systemname{}, a novel neurosymbolic framework. \systemname{} is proposed as a domain-specific grounded KBQA system that combines the generalization and reasoning capabilities of an LLM with an explainable context-gathering symbolic component. We demonstrate the utility of \systemname{} for two situations -- \emph{explainable context gathering} and \emph{explainable fact validation}.

\subsection{\systemname{} Components}
This framework expands the \emph{problem formulation-and-reasoning} paradigm by joining a neural generator to a symbolic component via a translator to improve the response’s explainability, and consequently trustworthiness.
\vspace{-1em}
\subsubsection{Symbolic Component}
Consisting of a domain-specific knowledge base and an inference agent, this component's role is to assemble explainable context, given an input query. Information stored in the knowledge base is represented in First-Order Logic and written in the form of Prolog rules and facts. This KB is also able to hold real-time information and user-specific preferences, and can be updated independently of the rest of the framework. The inference agent uses SWI-Prolog backward chaining to return the truth value conditioned on a Prolog query. 

\vspace{-1em}
\subsubsection{Neural Generator}
The neural generator is an LLM. The LLM serves to create a natural language response conditioned on a user query and context. In this work, we use GPT 3.5.
\vspace{-1em}
\subsubsection{Neural Translator}
We use this translator to facilitate the conversion from natural language (English) prose to Prolog and from Prolog to natural language prose. This is used to allow communication between our Neural Generator and our Symbolic Component. We use an LLM -- GPT 3.5 -- as our Neural Translator. 

\subsection{Explainable Context Gathering}
\begin{figure}
\scalebox{0.9}{\includegraphics[width=\textwidth]{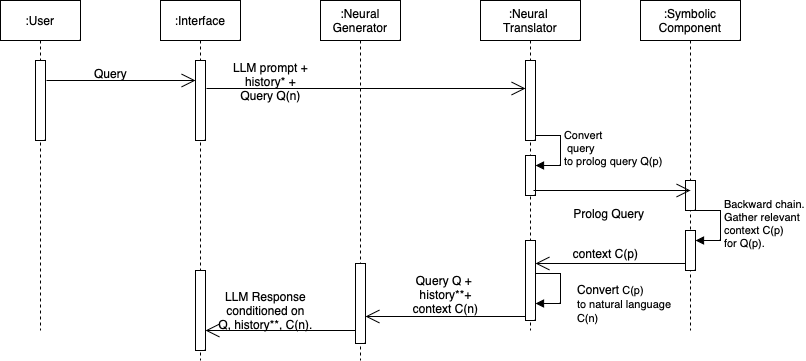}}
\caption{\small Flow of information diagram for \systemname{} in an explainable context gathering scenario. The system flow begins with the user query in natural language Q(n). The Neural Translator converts Q(n) to a Prolog query Q(p), which is fed to the inference engine. The inference engine runs a backward chain with the Q(p) as the goal and returns the context C(p) in the form of Prolog output, creating an goal tree in the process. The Neural Translator converts this context C(p) to a natural language context C(n). The Neural Generator is then fed with user query Q(n), context C(n), and a prompt P which informs the LLM to utilize C(n) and respond with reasoning. The final result is a Response R that is grounded by the human-interpretable, explainable, and accurate context used to generate it.} \label{fig1}
\end{figure}

In this configuration of \systemname{} (Fig \ref{fig1}), the symbolic component deterministically gathers and provides relevant context to the LLM for an explainable response generation. This setting is most useful where the reasoning behind the context grounds the LLM response, making it trustworthy. Scenarios such as providing answers that could be based on the user’s individual preferences, real-time information about the domain, or even information that does not exist officially but is built from experience are examples. 
For instance, in a community setting, a new member might want to use the swimming pool, but does not have much information about timings, busy hours, etc. A standalone LLM system, such as ChatGPT, would not have access to the relevant domain context and at best may provide a response such as this:
\begin{verbatim}
As an AI, I don't have real-time information about specific 
locations, including whether a pool is busy or not. You may 
want to check the pool's website or call them directly for 
the most up-to-date information on their current occupancy 
and busy times.
\end{verbatim}
\vspace{-2em}
\noindent \systemname{}, however, introduces relevant domain context about the pool through the symbolic component. As seen in Fig \ref{fig1}, the Translator converts the user query Q(n) to a prolog query Q(p).
\begin{verbatim}
Prompt: "you are a translator from english to prolog for ucsc. 
The global context is already ucsc. For every fact present in 
the user input, output it as a prolog goal. Keep the prolog goals as 
generic as possible. Return all goals as a LIST of strings. You 
do not need to mention ucsc in the prolog goals.
Example: 
User: UCSC has two libraries: McHenry and Engineering.
You: ['number_of_libraries(2).','library(mchenry).',
'library(engineering).']"

Input: Q(n): Is the pool busy?
Output: Q(p): status(pool, 1100, monday, Y).
\end{verbatim}
With Q(p) as the input, the Symbolic Component runs a backward chain, creating an implicit goal tree, shown in Fig \ref{goaltreepool}.

\begin{center}
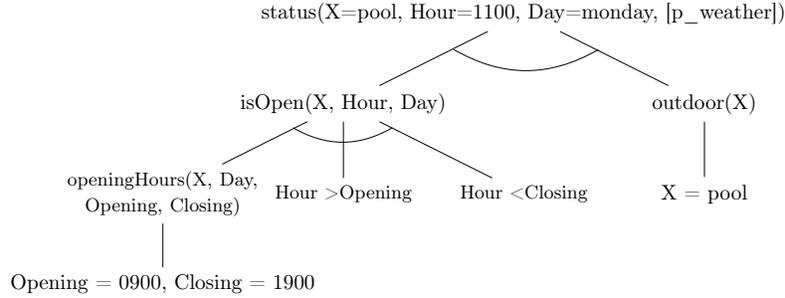

\scalebox{0.8}{
\begin{tikzpicture}[level distance=1.5cm,
  level 1/.style={sibling distance=6cm},
  level 2/.style={sibling distance=3cm}]
  \node(l1) {status(X=pool, Hour=1100, Day=monday, [p\_weather])}
    child { node(isOpen) {isOpen(X, Hour, Day)}
      child { node(opening)[text width=4cm,text centered] {\small{openingHours(X, Day, \\Opening, Closing)}}
        child { node {Opening = 0900, Closing = 1900} }
      }
      child { node(opening) {\small{Hour \textgreater Opening}}}
      child { node(closing) {\small{Hour \textless Closing}}}
    }
    child { node(outdoor) {outdoor(X)}
      child { node {X = pool} }
    };
    
    \draw[bend right] ($(isOpen.east)+(0,0.9)$) to ($(outdoor.west)+(-0.77,0.9)$);
    \draw[bend right] ($(opening.east)+(-2.1,1.1)$) to ($(closing.west)+(-1,1.1)$);
\end{tikzpicture}
}
    \captionof{figure}{\small Goal tree for the query \texttt{status(pool, 1100, monday, Y)}}
    \label{goaltreepool}
\end{center}

The Symbolic Component returns \textbf{Y}, the context C(p) or \texttt{p\_weather} in this case. The context can either be domain specific factors or they can be real time percepts. '\texttt{p\_weather}' is a real time percept and is replaced by its value. 
\begin{verbatim}
p_weather = sunny
\end{verbatim}
What should be noted is the reasoning behind returning '\texttt{p\_weather}' as context - it confirms the pool is open at the current time and also checks if it is an outdoor pool, which would make the weather affect its availability. This reasoning serves as an explanation for choosing the weather as a context that the Generator would use to create a response.\hfill\\\\
The Translator then converts C(p) to a natural language form C(n), a list of natural language contexts, by calling GPT 3.5 with the following prompt:
\begin{verbatim}
Prompt: "You are a translator from prolog to natural language. You will 
simply write out the prolog fact in natural language and append
the truth value to the end. 
Input: fact, {truth: 'TRUTH_VALUE'}
Output: fact (TRUTH_VALUE)"
\end{verbatim}
\begin{verbatim}
Input: C(p): p_weather = sunny 
Output: C(n): [`The weather is sunny']
\end{verbatim}
C(n) along with the original query Q(n) is fed in to the Generator, which then creates a grounded and explainable response R:
\begin{verbatim}
The weather is sunny. The pool may be busy on a sunny day as 
students may be looking to cool off and enjoy the nice weather.
It's always a good idea to check the pool's current capacity 
or availability before heading there.
\end{verbatim}

\subsection{Fact Validation}
\label{fact-checking}
\begin{center}
\begin{figure}
\scalebox{0.9}{\includegraphics[width=\textwidth]{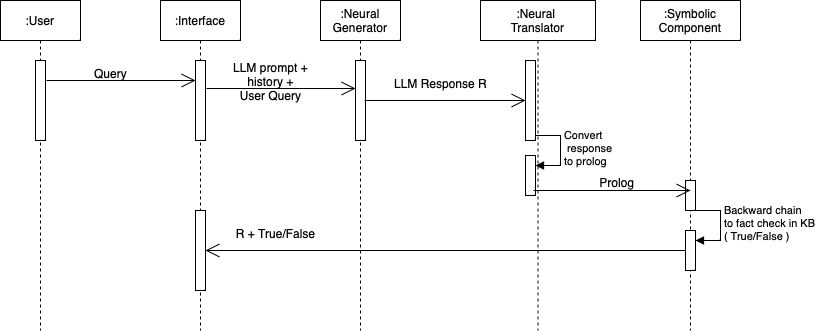}}
\caption{\small Flow of information diagram for \systemname{} in a fact checking scenario. The Neural Generator creates a response R based on the user query. This response R is fed into our Translator, which converts every fact in R into a list of Prolog queries. Every Prolog query is then be sent to our Symbolic Component for validation. For each query, the Symbolic Component returns True if that query is asserted by the KB, else False. These truth values are returned along with the original response R to the user.} \label{fig2}
\end{figure}
\end{center}

We can also use the Symbolic Component as a tool to validate LLM-generated responses to fact gathering questions. As seen in Figure \ref{fig2}, an LLM response can be asserted true if each fact in that response is individually asserted true. A goal tree complements every fact validation, making the response and its validation explainable and trustworthy. Note that since backward chaining can only deterministically assert a sentence, a False value should be interpreted as meaning either the sentence is factually false or the KB or the rule base is incomplete.

\vspace{-5em}
\section{Results}
\label{results}
We demonstrate the capability of \systemname{} in a specific domain. Here we focus on the use case of a KBQA system for a university campus, specifically UC Santa Cruz.
\vspace{-1em}
\subsection{Explainable Context Gathering}
We continue showing scenarios where explainable context gathering helps ground the response and improve its trustworthiness. We compare the output from ChatGPT 3.5 and the output from \systemname{}, complete with the reasoning steps. We also report the goal tree from the symbolic component. We return the context from the Symbolic Component in a variable \textbf{Y}.

\subsubsection{Query 1: Where can I study?} \hfill\\\\
\vspace{-0.5em}
Output from ChatGPT 3.5 (as accessed on March 2024) 
\begin{verbatim}
Where you can study depends on various factors such as 
your location, desired field of study, level of education, 
and personal preferences. Here are some general options:
Universities and Colleges: Traditional brick-and-mortar 
institutions offer a wide range of courses and degrees. 
Online Learning Platforms: There are numerous online platforms
such as Coursera, edX, Udemy, Khan Academy, and many others 
offering courses on various subjects...
\end{verbatim}
Output from Logic LLM 
\begin{verbatim}
You can study at the McHenry Library on campus. It 
is a quiet place with wifi available for students to use.
\end{verbatim}

\noindent Reasoning Steps taken
\begin{enumerate}
    \item Neural Translator (Natural Language Query to Prolog Query) 
    \begin{verbatim}
Output:     study_place(priyesh, Y)
    \end{verbatim}
    \vspace{-1em}
    \item Symbolic Component (Returns grounded context) \\\\
Goal Tree:  
\begin{center}
\scalebox{0.8}{
\begin{tikzpicture}[level distance=1cm,
  level 1/.style={sibling distance=6cm},
  level 2/.style={sibling distance=3cm}]
  \node {study\_place(X, [quiet, wifi])}
    child { node {student(X)} 
      child { node {X = priyesh} }
    };
\end{tikzpicture}
}
\end{center}
\begin{verbatim}
Output: prolog result: {`truth': True, `Y': [`quiet', `wifi']}
\end{verbatim}
The goal tree provides an explanation for the returned context - the study place should be quiet and should have wifi \textbf{if} priyesh is a student. \\

\item Neural Translator (Prolog context to Natural Language context)
\begin{verbatim}
Output: [`The place must be quiet', `The place must have wifi']
\end{verbatim}
\end{enumerate}

\subsubsection{Query 2: Can I drop classes?} \hfill\\\\
Output from ChatGPT 3.5 (as accessed on March 2024)
\begin{verbatim}
Whether you can drop classes depends on various factors such 
as your academic institution's policies, the timing within the
semester, and any specific requirements or consequences 
related to dropping classes. Typically, there are deadlines for
dropping classes without academic or financial penalties. These
deadlines vary from institution to institution and may differ 
based on the length of the semester or term.
\end{verbatim}
Output from Logic LLM 
\begin{verbatim}
Yes, you can drop classes, but please keep in mind that the
deadline to drop classes for this quarter is on the 29th of 
January. After this date, you will need to go through the
petition process to drop a class.
\end{verbatim}

\noindent Reasoning Steps taken

\begin{enumerate}
    \item Neural Translator (Natural Language Query to Prolog Query) \\
    \vspace{-1em}
    \begin{verbatim}
Output:     drop_classes(priyesh, 01, Y)
    \end{verbatim}
    \vspace{-1em}
    \item Symbolic Component (Returns grounded context) \\
Goal Tree:

\scalebox{0.8}{\begin{tikzpicture}[level distance=1.5cm,
  level 1/.style={sibling distance=4cm},
  level 2/.style={sibling distance=3cm}]
  \node {drop\_classes(X=priyesh, Month=01, ['29', 'jan'])}
    child { node(quarter) {quarter(winter, Start, End)}
      child { node {Start=01, End=03}}
    }
    child { node {Month $\geq$ Start}}
    child { node (end) {Month $\leq$ End}};
    \draw[bend right] ($(quarter.east)+(0.4,1)$) to ($(end.west)+(-1.3,1)$);

\end{tikzpicture}
}

    \begin{verbatim}
Output:  prolog result: {`truth': True, `Y': [`29', `jan']}
    \end{verbatim}
The goal tree here clearly shows the reasoning. The deadline to drop classes is 29 Jan if and only if it is the winter quarter, which is checked by looking at the current month of the year. \\

\item Neural Translator (Prolog context to Natural Language context)
\begin{verbatim}
Output:    [`The deadline is on the 29th of January']
\end{verbatim}
\end{enumerate}

\subsubsection{Query 3: Where should I dine today?} \hfill\\\\
Output from ChatGPT 3.5 (as accessed on March 2024)
\begin{verbatim}
Choosing where to dine depends on your preferences, budget, and
location. Here are some suggestions based on different types
of cuisine:
Italian: If you're in the mood for Italian cuisine, consider 
checking out a ristorante that serves authentic pasta dishes, 
wood-fired pizzas.
Mexican: For Mexican cuisine, look for a restaurant that 
serves traditional tacos, enchiladas.
\end{verbatim}
\noindent Output from Logic LLM 

\begin{verbatim}
You should dine at Stevenson. Stevenson College offers a variety
of dining options including a dining hall, cafe, and market.
\end{verbatim}

\noindent Reasoning Steps taken

\begin{enumerate}
    \item Neural Translator (Natural Language Query to Prolog Query) 
    \begin{verbatim}
Output:     dineAt(priyesh, 1300, monday, Y)
    \end{verbatim}
    \item Symbolic Component (Returns grounded context) \\
Goal Tree:

\scalebox{0.8}{\begin{tikzpicture}[level distance=1.5cm,
  level 1/.style={sibling distance=5cm},
  level 2/.style={sibling distance=2.5cm}]
  \node {dineAt(Name=priyesh, Hour=1300, Day=monday, Y)}
    child { node[text width=4cm,text centered] (prefer) {prefer\_cuisine (Name=priyesh, Cuisine)}
      child { node[text width=4cm,text centered] {prefer\_cuisine (priyesh, thai).}}
    }
    child { node {meal(Hour=1300, Type=lunch)}
        child { node(hour1) {Hour $>$ 1200}}
        child { node(hour2) {Hour $<$ 1500}}
    }
    child { node[text width=4cm,text centered] (cuisine) {cuisineAt(Day=monday, Type, Cuisine, DiningHall)}
        child { node[text width=4cm,text centered] {cuisineAt(monday, lunch, thai, stevenson).}}
    };
    \draw[bend right] ($(prefer.east)+(1.3,1)$) to ($(cuisine.west)+(-1.2,1)$);
    \draw[bend right] ($(hour1.east)+(-0.3,1)$) to ($(hour2.west)+(0.25,1)$);

\end{tikzpicture}
}
\begin{verbatim}
Output:  prolog result: {'truth': True, 'Y': 'stevenson'}
\end{verbatim}
\item Neural Translator (Prolog context to Natural Language context)
\begin{verbatim}
Output:    ['You should dine at Stevenson']
\end{verbatim}
The reasoning here, seen in the goal tree above, takes in the user's culinary preferences into account to provide an accurate context - in this case which dining hall to eat at.
\end{enumerate}

\subsection{Fact Validation}
We also demonstrate \systemname{}'s flexibility by employing it to improve explainability in another way -- fact checking. In these scenarios, one can use the framework to determine the truth value of statements or model outputs. We show an example of our system validating factual claims in a GPT-generated response against our UCSC Knowledge Base. 

\subsubsection{Query: How many dining halls does UCSC have?} \hfill\\\\
Output from ChatGPT 3.5 (as accessed on March 2024)

\begin{verbatim}
UCSC has three dining halls: College Nine/Ten Dining Hall,
Cowell/Stevenson Dining Hall, and Crown/Merrill Dining Hall.
\end{verbatim}
Output from \systemname{}

\begin{verbatim}
UCSC has three dining halls: College Nine/Ten Dining Hall,
Cowell/Stevenson Dining Hall, and Crown/Merrill Dining Hall.
\end{verbatim}
\begin{verbatim}
validation: 
number_of_dining_halls(3) (False)
dining_hall(college_nine_ten) (True)
dining_hall(cowell_stevenson) (False)
dining_hall(crown_merrill) (True)
\end{verbatim}

\vspace{-1em}
\section{Discussion and Conclusion}
In this work we introduced \systemname{}, a novel neurosymbolic framework for domain-specific KBQA systems, and demonstrated the importance of explainable context gathering and fact validation for a variety of scenarios in Section \ref{results} above. 

\systemname{} is able to provide correct and reasonably sound `opinions' and  `verified factual responses' due to either the curated context or the fact validation  respectively. The symbolic component in \systemname{} produces an explainable chain of reasoning to contextualize a query, as is the case in the context gathering task, or to fact check a statement, as is the case in fact validation, while leveraging the inherently uninterpretable, powerful generative abilities of an LLM. This makes the LLM response, that is conditioned on this context, trustworthy and explainable. As seen in Query 3, by providing context built on the users preferences, this also allows us to personalize the response for a user. 
This approach holds significant promise for the field of explainable AI; especially given the fact our framework does not need any LLM retraining

One clear limitation is having an incomplete KB, which then will make \systemname{} return an inaccurate answer. One way to mitigate this is to manually update the KB every time, but that's not feasible. Hence, we plan to implement user-interactive updates to the KB. In other words, if the user notices discrepancies in the goal tree created, they will be able to write a natural language text that \systemname{} will utilize to update the KB. \systemname{} can then be personalized to individual users -- each person having a local knowledge base that is continually updated with user preferences, new facts, etc. With this added functionality, the system can be adapted to new domains with ease, and scaled up to handle more use cases within a given domain.

Another limitation of \systemname{} is that while the goal trees are accurate to the internal workings of the inference engine, they are not currently returned by the symbolic component. We plan to implement a user-accessible visualization of the step-by-step backward chaining in the symbolic component -- further improving the explainability of the system. 

LLMs are powerful, entering many domains in everyday life, they are becoming responsible for human-level high impact decisions, they need to be trustworthy, our \systemname{} system is a step towards trustworthy, explainable by design.


\section{Acknowledgements}
Funding for this work was provided by Underwriters Laboratories Inc. through
the Center for Advancing Safety of Machine Intelligence.
\
\bibliographystyle{splncs04}
\bibliography{mybibliography}
%
%




%
\end{document}